\documentclass[letterpaper, 10 pt, conference]{ieeeconf}  % Comment this line out if you need a4paper

\IEEEoverridecommandlockouts                              % This command is only needed if 
\newcommand{\argmax}{\mathop{\mathrm{arg\,max}}}

\usepackage[utf8]{inputenc}
\usepackage[T1]{fontenc}
\usepackage[disable,textsize=scriptsize]{todonotes}
\usepackage{url}
\usepackage{lipsum}
\usepackage[separate-uncertainty = true]{siunitx}
\usepackage{amsmath} % assumes amsmath package installed
\usepackage{amsfonts}
\usepackage{amssymb}  % assumes amsmath package installed
\usepackage{bm}

\newcommand{\insertref}[1]{\todo[color=green!40]{#1}}

\title{\LARGE \bf
Learning Skill-based Industrial Robot Tasks with User Priors
}

\author{Matthias Mayr$^{1}$, Carl Hvarfner$^{1}$, Konstantinos Chatzilygeroudis$^{2}$, Luigi Nardi$^{1,3}$ and Volker Krueger$^{1}$% <-this % stops a space
	\thanks{$^{1}$Department of Computer Science, Faculty of Engineering (LTH), Lund University, SE~221~00 Lund, Sweden. E-mail: <firstname>.<lastname>@cs.lth.se.
	}%
	\thanks{$^2$Computer Engineering and Informatics Department (CEID), University of Patras, Greece. E-mail: costashatz@upatras.gr.}
	\thanks{$^{3}$
	Department of Computer Science and Electrical Engineering, Stanford University, CA 94305, USA. E-mail: lnardi@stanford.edu.}
}

\begin{document}

\maketitle
\thispagestyle{empty}
\pagestyle{empty}

%%%%%%%%%%%%%%%%%%%%%%%%%%%%%%%%%%%%%%%%%%%%%%%%%%%%%%%%%%%%%%%%%%%%%%%%%%%%%%%%
\begin{abstract}
Robot skills systems are meant to reduce robot setup time for new manufacturing tasks. Yet, for dexterous, contact-rich tasks, it is often difficult to find the right skill parameters. One strategy is to learn these parameters by allowing the robot system to learn directly on the task. 
For a learning problem, a robot operator can typically specify the type and range of values of the parameters. Nevertheless, given their prior experience, robot operators should be able to help the learning process further by providing educated guesses about where in the parameter space potential optimal solutions could be found.
Interestingly, such prior knowledge is not exploited in current robot learning frameworks.
We introduce an approach that combines user priors and Bayesian optimization to allow fast optimization of robot industrial tasks at robot deployment time. We evaluate our method on three tasks that are learned in simulation as well as on two tasks that are learned directly on a real robot system. Additionally, we transfer knowledge from the corresponding simulation tasks by automatically constructing priors from well-performing configurations for learning on the real system.
To handle potentially contradicting task objectives, the tasks are modeled as multi-objective problems.
Our results show that operator priors, both user-specified and transferred, vastly accelerate the discovery of rich Pareto fronts, and typically produce final performance far superior to proposed baselines.\end{abstract}

%%%%%%%%%%%%%%%%%%%%%%%%%%%%%%%%%%%%%%%%%%%%%%%%%%%%%%%%%%%%%%%%%%%%%%%%%%%%%%%%
\section{INTRODUCTION}
In modern manufacturing settings, the setup of a robot system for a new task should be fast and easy. 
At the same time, to assure safety of equipment and workers it is important that robot behavior is always predictable and explainable. %, i.e., to know \emph{what} the robot is doing \emph{when} and \emph{why}.

One way to combine all these requirements is a system based on modular and explainable \emph{robot skills}~\cite{krueger16ieee}. Robot skills (or just skills) are semantically defined parametric actions where parameters have to be chosen based on the task at hand through planning, sensing and knowledge integration. 
For contact-rich tasks, however, it can still be very challenging to find well-functioning skill parameter values, as even human operators may encounter difficulties identifying a successful and robust parameter set~\cite{mayr22iros}. One solution is to allow a robot system to find these parameters through reinforcement learning (RL) directly on the task.
A recent approach~\cite{mayr21iros} to RL in the industrial context suggests to model explainable policies with behavior trees (BTs) and a motion generator (MG)~\cite{rovida182iicirsi}, and to optimize these through efficient policy learning and domain randomization within a digital twin.
\begin{figure}[tpb]
	{
		%\setlength{\fboxrule}{0pt}
		%\framebox{
		%\parbox{3in}{
		\begin{center}
		    \includegraphics[trim=0 60 0 55, width=\columnwidth]{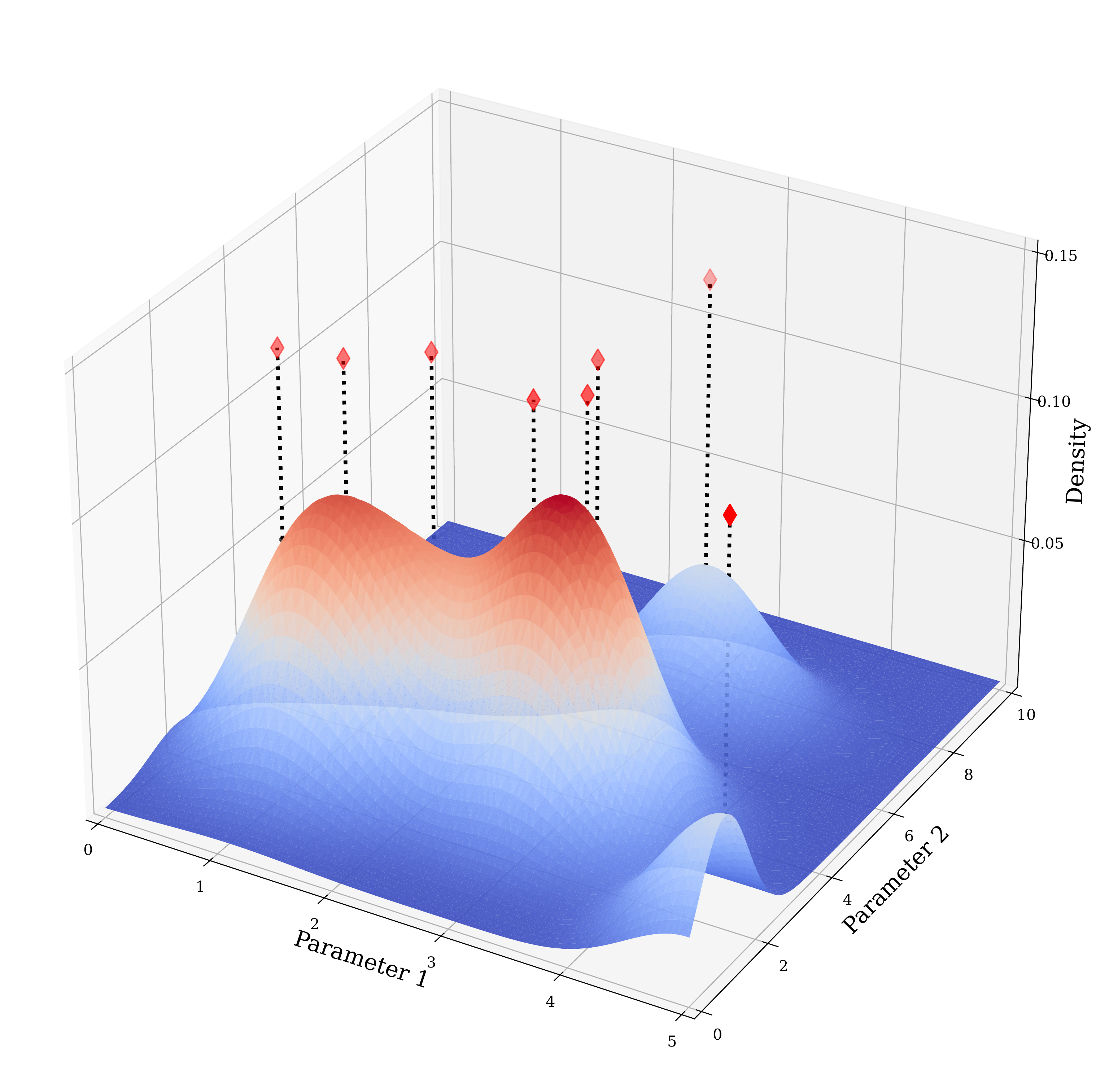}    
		\end{center}
		%}}
	}
	\caption{The visualization of two dimensions of a multimodal prior formed by results from learning in simulation that can be used as a prior for learning on the real robot. Well-performing configurations in simulation (red diamonds) are used to construct a probability density over optimal parameter settings, enabling accelerated learning on the real task through Bayesian optimization.}
	\label{fig:prior}
\end{figure}
However, beyond the straight learning problem, there are two important aspects to consider: 
%\begin{enumerate}
%    \item 

(1) Learning often needs to balance various key performance indicators (KPIs) such as robot speed, safety or the need to minimize interaction forces with manufacturing parts. While~\cite{mayr21iros} is able to handle only single-objective learning, we argue that many tasks are best described as multi-objective learning problems where the outcome is a variety of policies for different trade-offs between the objectives~\cite{mayr22iros}. An operator can then choose a solution with the desired properties.
%    \item

(2) The learning problem can be reduced by constraining the parameter space within which the RL approach is searching for suitable skill parameters. Given some prior experience, a robot operator can typically not only help to constrain the search space, but further accelerate the learning process by providing guesses on where in the parameter space optimal solutions may be found, and have these regions be emphasized throughout the learning process. These guesses could be based on the operator's intuition or experience, or be generated from previously utilized policies on similar tasks through transfer learning. Fig.~\ref{fig:prior} visualizes how guesses regarding good configurations can be represented through a \textit{probability density over optimal parameter settings}. 
    In fact, by utilizing knowledge in the form of a probability density over the optimum, the search can focus on promising areas of the search space without explicitly restricting the search space to these regions.
%\end{enumerate}

\noindent
With this paper we make the following contributions:
\begin{enumerate}
    \item We introduce an approach to incorporate parameter priors in the form of probability densities for the optimal configuration, in conjunction with multi-objective Bayesian optimization, into the learning process of industrial robot tasks.
    \item We assess the performance of our method and evaluate the influence of well-placed and misleading priors on various tasks.
    \item We show priors learned in simulation can enable accelerated optimization on the real system, without the requirement of explicit operator knowledge.
\end{enumerate}

\section{Related Work}
\subsection{Reinforcement Learning with Robot Systems}

Reinforcement Learning (RL)~\cite{sutton1998reinforcement}, and especially Direct Policy Search (PS) methods~\cite{chatzilygeroudis2019survey,deisenroth_survey_2013}, have been successful in robotics applications as they can be applied in high-dimensional continuous state-action problems. In order to apply PS methods successfully in robotics applications, one must do one (or a combination) of the following: (a) provide prior structure in the policy, (b) learn models of the dynamics or the expected return, and/or (c) use prior information about the search space~\cite{chatzilygeroudis2019survey}.

The type of the policy structure plays an important role for the effectiveness of learning in practical robotics applications. However, there is always a tradeoff between having a representation that is expressive enough (e.g. large neural networks), and one that provides a space that is efficiently searchable~\cite{chatzilygeroudis2019survey}. Another important property is choosing the level on which the policy interacts with the robot (e.g. task-space vs joint-space); it has been shown that it can strongly influence the learning speed and the quality of the obtained solutions~\cite{varin2019policy, martin2019imp}.

Traditionally, the robotic controllers (or policies) are hand-designed; either via an analytic model-based approach (e.g. inverse dynamics controllers)~\cite{peters2008learning} or as more general finite state machines (FSMs)~\cite{calandra2016bayesian}. These hand-designed policies usually come with a small amount of parameters (thus efficiently searchable), but might need to be re-designed when changing task or robot. In principle, most of the hand-designed policies are easily intepretable and we can infer why the robot is choosing a specific action.
% Recently, policies based on Behavior Trees have been proposed~\cite{mayr21iros,mayr22iros,colledanchise17ac} that are a form of 

The most popular way of defining a policy in the RL literature is as a function approximator (e.g. a neural network)~\cite{stulp2013robot,chatzilygeroudis2019survey,deisenroth_survey_2013}. In this case, policies can be very expressive and task-agnostic, which means that the same policies can be re-used on different tasks or robots without substantial changes. The commonly used policy representations for learning systems include radial basis function networks~\cite{deisenroth13r}, dynamical movement primitives~\cite{ijspeert2013dynamical, ude2010task} and feed-forward neural networks~\cite{deisenroth13r,chatzilygeroudis172iicirsi}. In recent years, in RL settings, deep artificial neural networks (ANNs) have become the default policy type~\cite{arulkumaran2017brief,chatzilygeroudis2019survey}. ANN policies enable us to easily increase the expressiveness and generality of the policy, but can make optimization difficult due of the large number of parameters. In contrast to the previous policy types, this type of policy is harder to interpret, and it is generally a difficult task to know why the robot is choosing a specific reaction to an environmental change.
Therefore,~\cite{mayr21iros} and~\cite{mayr22iros} suggest to learn interpretable policies based on BTs and a MG~\cite{rovida182iicirsi} that are well suited for the requirements of an industrial environment.

\subsection{Meta-learning for Bayesian Optimization}
For the optimization of black-box functions, Bayesian Optimization (BO) constitutes a sample-efficient \cite{frazier2018tutorial} choice across multiple fields, including machine learning~\cite{snoek-nips12a}, robotics~\cite{calandra-lion14a}, and hardware design~\cite{nardi18hypermapper}. In BO, there are several means of injecting prior knowledge, the most common of which is through the choice of the Gaussian Process kernel. However, several approaches have been proposed to explicitly bias or direct the optimization, based on accumulated data or knowledge from previous tasks.

Transfer learning approaches make use of data obtained from previous experiments to guide current ones. Feurer et. al. \cite{feurer2018practical} propose to combine surrogate models from previous experiments, and use the combined surrogate when performing a new task. Perrone et. al. \cite{perrone-neurips18a}, restrict the search space of the new task to some convex region based on optima found on previous tasks, excluding suboptimal regions in the outer edges of previous search spaces.

Injecting explicit prior distributions over the location of an optimum is an emerging topic in BO. In these cases, the user explicitly defines a prior probability distribution $\pi(\bm{x})$ that encodes their belief on where the optimum is likely to be located. Souza et. al. propose \emph{BOPrO}~\cite{souza2021bayesian}, which combines $\pi(\bm{x})$ with a data-driven model into a pseudo-posterior. From the pseudo-posterior, configurations are selected using the Expected Improvement (EI) acquisition function. Hvarfner et. al. \cite{hvarfner2022pibo} propose $\pi\text{BO}$, which weights the acquisition function by $\pi(\bm{x})$, and decays the prior's influence over time. Consequently, it retains conventional convergence rates \cite{JMLR:v12:bull11a} for any choice of $\pi(\bm{x})$ when used in conjunction with EI.

\section{Approach}
In order to learn robot tasks, we utilize two main components: 1) \emph{SkiROS2}~\cite{rovida2017skiros,krueger16ieee} is a skill-based system for \emph{ROS}. It provides a world model (digital twin) and a skill representation based on behavior trees (BT), and has an integrated task planner. 2) An RL framework that integrates optimizers and provides a simulation as well as reward calculation~\cite{mayr21iros,mayr22iros}.

When setting up the system for a new task, the operator typically specifies a high-level goal using the \emph{Planning Domain Definition Language} (PDDL). Once the planner finds a valid sequence of skills, the learnable parameters in the skills are automatically identified.
The operator can state lower and upper bounds
for the parameters to be learned. A more detailed description can be found in~\cite{mayr22iros}. In this work, we additionally allow for specification of a unimodal or multimodal prior for the optimum. Therefore, expert knowledge and  previous experiences can be actively integrated into the learning process.

\subsection{Skill Representation}
We adopt the skill definition from~\cite{krueger16ieee,rovida2017skiros} that defines a skill as an ability to change the world state. To support task planning, a skill has a set of pre-conditions that must be satisfied before the execution is started and post-conditions that state and verify the effects. Skills usually model instructions from standard operation procedures (SOPs),  such as \emph{pick<object>}, \emph{insert<object>}, \emph{press<object>}, etc.

Our parametric skills can utilize the world model to retrieve knowledge and are implemented with BTs.
A BT~\cite{colledanchise17ac}, is a plan representation and execution tool that is used in many areas including computer games and robotics~\cite{colledanchise17ac,iovino2020survey}. As in~\cite{rovida172iicirsi,marzinotto142iicrai}, we define it as a directed acyclic graph with nodes and edges. It consists of \emph{control flow nodes} or \emph{processors} that link \emph{execution nodes}. Three common types of \emph{control flow nodes} are 1) \emph{sequence} (logical \emph{AND}), 2) \emph{selector} (logical \emph{OR}) and 3) \emph{parallel}. A BT always has one initial node with no parents, called \emph{Root node}. During the execution of a BT, a periodic \emph{tick signal} is injected into the \emph{Root node}. The signal is routed according to the \emph{control flow nodes} and the return statements of the children.
The leaves of the BT are the \emph{execution nodes} that execute one cycle and output one of the three signals when being ticked: \emph{success}, \emph{failure} or \emph{running}.
Execution nodes subdivide into 1)~\emph{action} and 2)~\emph{condition} nodes. An action node performs its operation iteratively at every tick and returns \emph{running} while it is not done, and \emph{success} or \emph{failure} otherwise. A condition node performs an atomic operation and can only return \emph{success} or \emph{failure}, but never \emph{running}. 
One significant difference between BTs and FSMs is that BTs implement a two-way control flow like function calls in programming languages. In contrast, classical FSMs implement a one-way control flow similar to \emph{GOTO} statements, which often becomes challenging to scale and maintain.

For modeling parametric movements, our movement skills use a MG in combinations with BTs~\cite{rovida182iicirsi}. This formulation is a type of trajectory-based policy structure that explicitly operates in end-effector space. The advantages of such movement skills are that they are modular, interpretable and allow for an easy adaption to environmental changes, e.g. if objects are relocated. In line with~\cite{rovida182iicirsi}, we require a compliant robot controller that operates in end-effector space and utilize the same Cartesian impedance controller as in~\cite{mayr22iros}. 

\subsection{Policy Optimization} \label{sec:policy}
\label{sec:optimization}
In order to optimize for policy parameters, we adopt the policy search formulation in~\cite{deisenroth13r,chatzilygeroudis2019survey,chatzilygeroudis172iicirsi}. %
We formulate a dynamical system of the form:
\begin{equation}
\label{formula:dynamics}
\mathbf{x}_{t+1}=\mathbf{x}_{t}+M(\mathbf{x}_{t}, \mathbf{u}_{t}, \boldsymbol{\phi}_R)
%, \boldsymbol{\phi}_{K}
,
\end{equation}
with continuous-valued states $\mathbf{x} \in \mathbb{R}^E$ and actions $\mathbf{u} \in \mathbb{R}^U$.
The transition dynamics are modeled by a simulation of the robot and the environment $M(\mathbf{x}_t, \mathbf{u}_t, \boldsymbol{\phi}_R)$. They are influenced by the domain randomization parameters $\boldsymbol{\phi}_R$.

The goal is to find a policy $\pi, \mathbf{u} = \pi(\mathbf{x}|\boldsymbol{\theta})$ with policy parameters $\boldsymbol{\theta}$ such that we maximize  the expected long-term reward when executing the policy for $T$ time steps:%, given a distribution of the domain parameters $\theta$:
\begin{equation}
\label{formula:long_term_reward}
J(\boldsymbol{\theta}) =\mathbb{E} \left[\sum_{t=1}^{T} r(\mathbf{x}_{t},\mathbf{u}_{t}) | \boldsymbol{\theta}\right]
%P(\boldsymbol{\theta})
,
\end{equation}
where $r(\mathbf{x}_{t},\mathbf{u}_{t})$ is the immediate reward for being in state $\mathbf{x}$ and executing action $\mathbf{u}$ at time step $t$.
The discrete switching of branches in the BT and most skills are not differentiable. Therefore, we frame the optimization in Eq.~\eqref{formula:long_term_reward} as a black-box function and pursue the maximization of the reward function $J(\mathbf{\theta})$ only by using measurements of the function. The optimal reward function to solve the task is generally unknown, and a combination of reward functions is usually used. In the RL literature, this is usually done with a weighted average, that is, $r(\mathbf{x}_{t},\mathbf{u}_{t})=\sum_iw_ir_i(\mathbf{x}_{t},\mathbf{u}_{t})$. In this paper, we choose not to use a weighted average of reward functions that represent different objectives (as the optimal combination of weights cannot always be found~\cite{kaushik2018multi}), but optimize for all objectives concurrently (Sec.~\ref{sec:multi-objective}) using Bayesian Optimization.

\subsection{Bayesian Optimization} \label{sec:bo}
As mentioned in Section~\ref{sec:policy}. we view the of optimization of our policy as an unknown black-box optimization problem. In this setting, information about the objective function $f$ can only be extracted through the potentially noisy output $y$ yielded by an given input $\bm{x}$. We wish to find $\bm{x}^* \in \argmax _{\bm{x}\in \mathcal{X}} f(\bm{x})$ for some bounded, $D$-dimensional input space $\mathcal{X}$. As the function $f$ is typically expensive in some resource of interest, one wishes to optimize $f$ with a low total number of evaluations.

To solve the aforementioned black-box optimization problem, we employ BO. It aims to find $\bm{x}^*$ by sequentially selecting new design points $\{\bm{x}_i\}_{i=1}^N$ through some measure
of utility, then receiving their corresponding output $\{y_i\}_{i=1}^N$, for some maximal number of iterations $N$. This is achieved through the use of a probabilistic surrogate model $p(f|\mathcal{D}_n)$, which uses all available observations $\mathcal{D}_n = \{\bm{x}_i, y_i\}_{i=1}^n$ at a given iteration $n$ to emulate the objective $f$. After obtaining an initial number of observations through some space-filling design (Design of Experiments, DoE). BO uses the aforementioned utility measure, commonly called an acquisition function, to decide on subsequent queries. A query is selected $\bm{x}_{n+1}$ by considering a trade-off between uncertain regions (exploration) and regions of high predicted value (exploitation) under $p(f)$. After evaluation, the observation $y_{n+1}$ is obtained, and the surrogate model is updated. The most commonly used acquisition function is Expected Improvement (EI)~\cite{jones-jgo98a, JMLR:v12:bull11a}, which is defined as
\begin{equation} 
\bm{x}_{n+1} \in \argmax_{\bm{x}\in\mathcal{X}} \mathbb{E}_y\left[[(y^*_n - y(\bm{x})]^+\right]
\end{equation}
where $y_n^*$ is the best obtained (noisy) output at iteration $n$.  EI is simple to implement, and can be computed closed-form.

For tasks with a substantial level of noise, such as robot learning, the consideration of noise in the objective $f$ is of particular importance~\cite{calandra2015bayesian}. EI can potentially struggle in such noisy settings~\cite{Vazquez_2008, letham2018noisy, gramacyei} due to its consideration of the improvement of a noisy observation. As such, we utilize a noisy-robust version, called Noisy EI (NEI)~\cite{letham2018noisy}, defined as
\begin{equation}\label{eq:nei}
    \bm{x}_{n+1} \in \argmax_{\bm{x}\in\mathcal{X}}
    \mathbb{E}_f\left[
    [(f^*_n - f(\bm{x})]^+\right]
\end{equation}
which, despite similarities to EI, requires approximation through Monte Carlo by sampling latent function values at each prior observed location. Through its consideration of the noiseless optimum $f^*$ as opposed to  $y^*$, NEI, yields desired robustness to noise and converges to EI in a noiseless setting. For our experiments, we use the \emph{HyperMapper} \cite{nardi18hypermapper, nardi2017algorithmic} framework, and the NEI acquisition function introduced in~Eq.\eqref{eq:nei},
modified for a multi-objective setting. The modification is covered exhaustively in Section \ref{sec:multi-objective}.

\subsection{Multi-objective Optimization} \label{sec:multi-objective}
In the multi-objective optimization setting, we consider a set of $K$ objectives $\bm{f} = (f_1, \ldots, f_K)$, all defined over the same $D$-dimensional input space $\mathcal{X}$ and observed through a noisy output $\bm{y} = y_1, \ldots, y_K$. Our goal is to find the set of points that are not dominated by any other point in $\mathcal{X}$. For a pair of inputs $\bm{x}$ and $\bm{x}'$ with corresponding multi-objetive outputs $\bm{y}$ and $\bm{y}'$, $\bm{x'}$ is said to dominate $\bm{x}$ if $y_k' \geq y_k, \forall k \in \{1, \ldots K\}$, i.e. $\bm{y}'$ is superior in every objective. The set of non-dominated points, known as the Pareto frontier, is in turn expressed as $\Gamma = \{\bm{x} \in \mathcal{X}: \nexists \bm{x}' \; \text{s.t.} \; \bm{x}' \prec \bm{x} \}$, where $\prec$ is the domination relation. $\Gamma$ thus contains the set of maximally desired points, with various trade-offs in the objectives.

For our experiments, we use the random scalarizations approach proposed by Paria et. al. \cite{Paria2019AFF}, which computes the acquisition function across objectives as 
\begin{equation}
      \alpha(\bm{x}, \bm{\lambda}) = \sum_{k=1}^K\lambda_k\alpha_k(\bm{x}), \quad \sum_{k=1}^K \lambda_k = 1
\end{equation}
on the $K$ objective-wise acquisition functions $\{\alpha_k\}_{k=1}^K$ and the scalarization $\bm{\lambda}$, sampled from a Dirichlet distribution. To quantify the quality of the obtained Pareto front, we use the Hypervolume Indicator (HV) \cite{CAO201560} metric. HV computes the volume spanned by the Pareto-optimal observations $\{\bm{y}_p\}_{p=1}^{|\Gamma|}$ from some reference point $\bm{r}$ as $\mathcal{HV}(\Gamma, \bm{r}) = \lambda_K\left(\cup_p [\bm{y}_p, \bm{r}]\right)$ where $[\bm{y}_p, \bm{r}]$ denotes the hyperrectangle bounded by vertices $\bm{y}_p$ and $\bm{r}$, and $\lambda_K$ is the $K$-dimensional Lebesgue measure. 

\subsection{Priors for the Optimum}
For our experiments, we use the \emph{HyperMapper} implementation of $\pi\text{BO}$ \cite{hvarfner2022pibo}, combined with the NEI acquisition function. Moreover, we utilize the prior for sampling during DoE. Critically, $\pi(\bm{x})$ is defined on the input space $\mathcal{X}$. As such, the user defines one prior jointly over all objectives, so that the prior emphasizes regions that are believed to contribute to the Pareto front, with no emphasis towards any particular objective.

The tasks use two types of priors over the optimum:\\
\textbf{Simulation}:
    We consider Gaussian densities for each parameter. These are set once, prior to conducting the experiments by an expert operator. The priors are left untouched for the whole duration of the experiments to avoid bias.\\
\textbf{Real system}: In addition to the operator priors, we use $\pi\text{BO}$ in a transfer learning setting, as we form the prior based on previous data. The Pareto front designs obtained from simulation are used to form a Gaussian kernel density estimator (KDE)~\cite{parzen1962estimation}. KDE places a Gaussian density on each point on $\Gamma$, which enables the automatic construction of multimodal distributions, exemplified by Figure \ref{fig:prior}. The obtained Pareto front from simulation then serves as a starting point for learning in the real system, without involving the operator.

\section{Experiments}
\begin{figure}[tpb]
	{
		\setlength{\fboxrule}{0pt}
		\framebox{\parbox{3in}{
		\includegraphics[width=0.95\columnwidth]{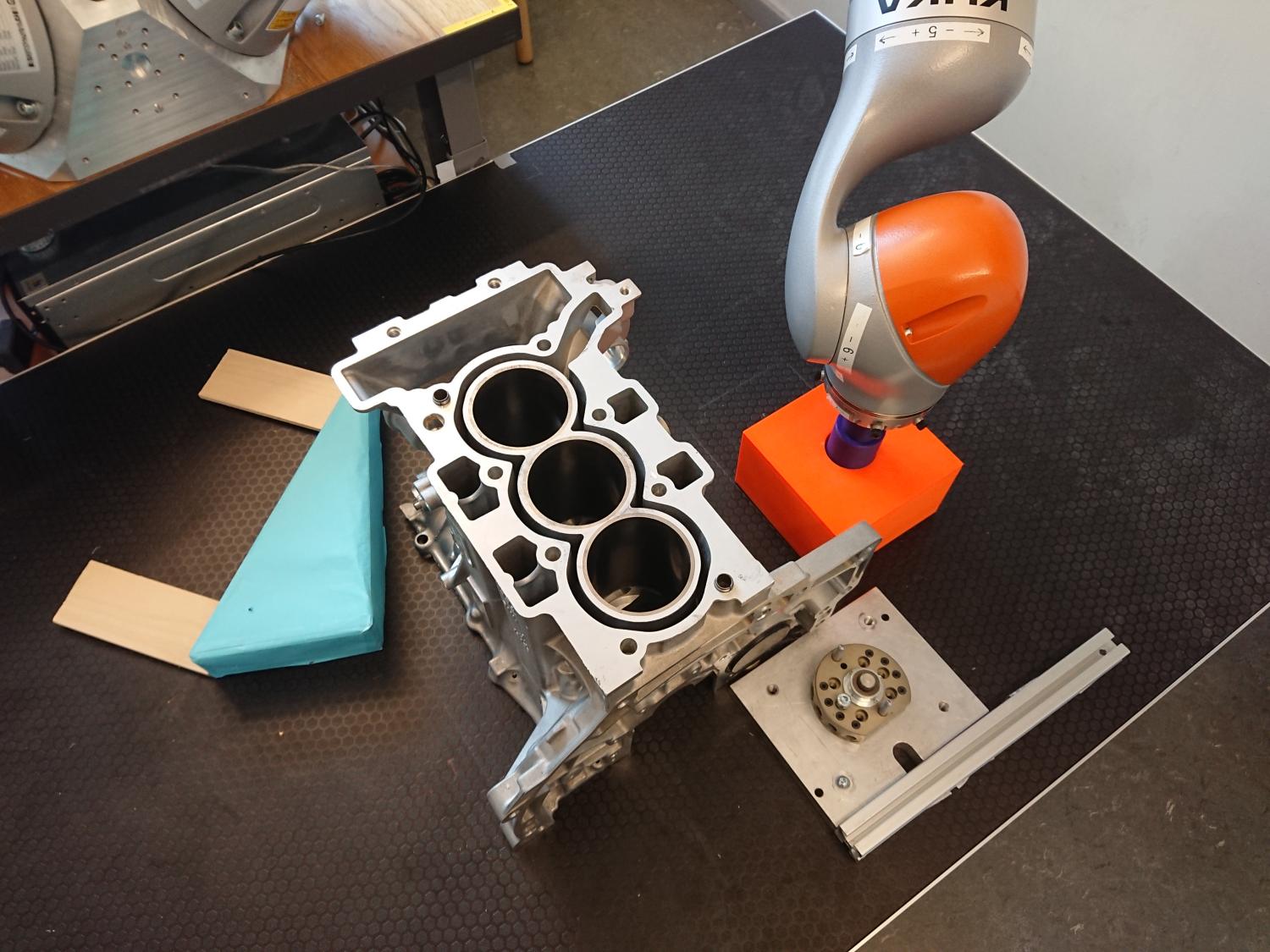}}}
	}
	\caption{The experimental setup for the tasks. The orange block is used for the peg insertion. The engine is the obstacle that must be avoided with the use of movement skills when transitioning from one side to the other. The push task requires the blue object to be pushed from its current pose to the corner between the box and the fixture when the engine is not at the workstation.}
	\label{fig:exp-setup}
\end{figure}

\begin{figure}[tpb]
	{
		\setlength{\fboxrule}{0pt}
		\framebox{\parbox{3in}{
		\begin{center}
		    \includegraphics[width=0.85\columnwidth]{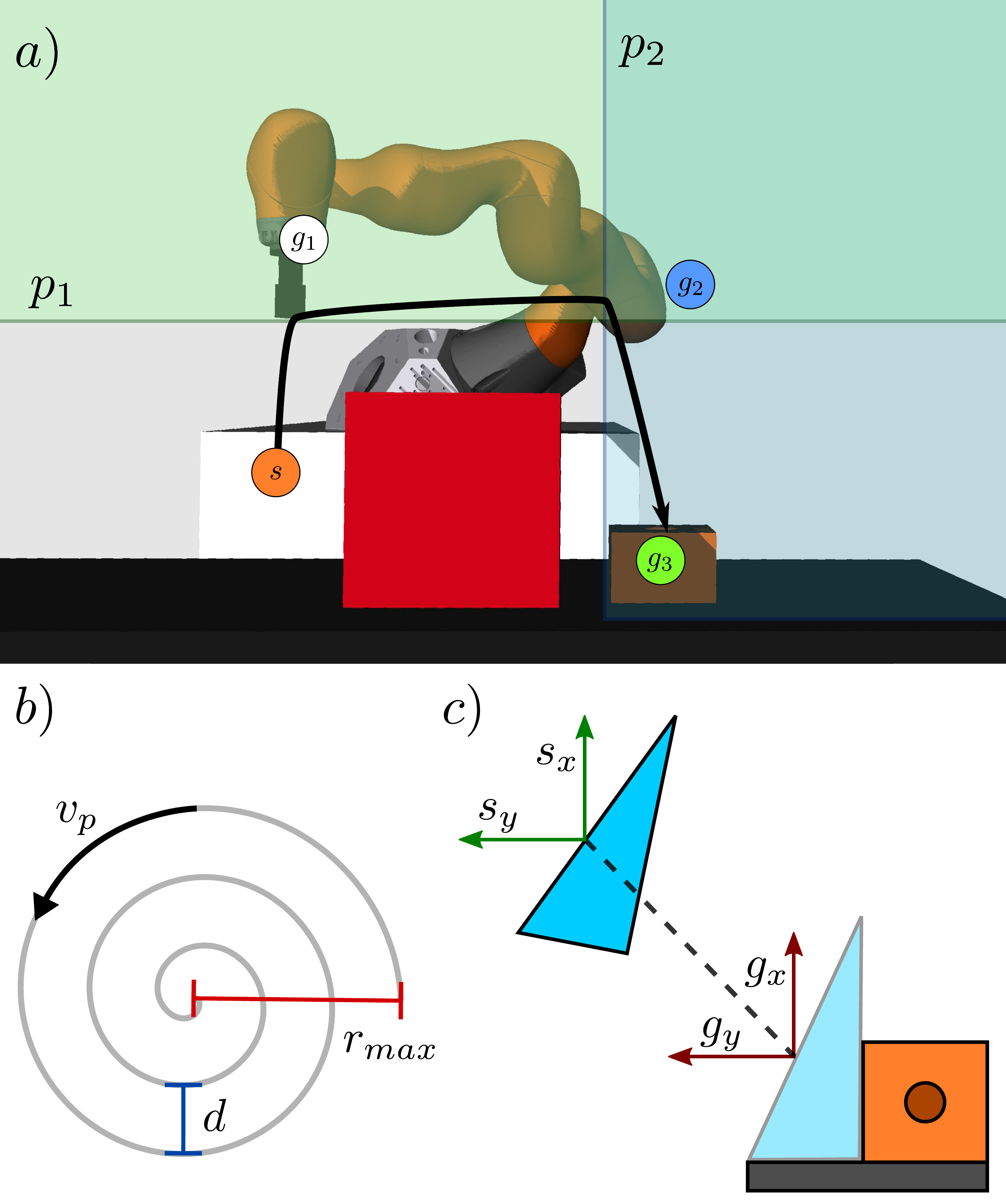}
		\end{center}    
	    }}
	}
	\caption{A depiction of the learnable parameters of the different tasks. a) The setup of the obstacle avoidance task with the parametric goal points \(g_1\) and \(g_2\) and the adjustable thresholds \(p_1\) and \(p_2\) in one possible motion configuration. b) The spiral of the search motion for the peg insertion is defined by the pitch \(d\), the maximal \(r_{max}\) and the path velocity \(v_p\). In addition, a downward force is set. c) The learnable offsets for the start and goal location of the push task are shown.}
	\label{fig:learnable}
\end{figure}

We evaluate the influence of priors for the optimum on the learning process for three different tasks. Their setup on the workstation is shown in Fig.~\ref{fig:exp-setup};  Fig.~\ref{fig:learnable} shows the learnable parameters. One of the tasks is a contact-free movement from one side of an object (e.g., the engine block in Fig.~\ref{fig:exp-setup}) to the other side. The other two tasks are contact-rich manipulations where a peg needs to be inserted into a hole and an object with an uneven weight distribution needs to be pushed to another location, respectively. All tasks have in common that they are solved with existing skills that use BTs and a MG to actuate the robot and where some skill parameters need to be found via learning. For each task, a learning scenario configuration file describes attributes such as the robot system to use, the configured reward functions and the learnable skill parameters with their bounds and optionally their priors.

All learning problems are defined as multi-objective problems where one objective assesses the performance and speed of solving the task and the other one is either a safety metric, defined by distance between the robot and a fragile item when passing it, or an impact metric, which considers the interactions forces of the robot and the work pieces. None of the tasks have a single best policy, and since they balance trade-offs between the competing objectives (KPIs), it is up to an operator to decide which one to use as a final policy. We utilize the HV defined by the Pareto-optimal points~\insertref{MO metrics} to measure how much of the solution space is covered. We locate the reference point $\bm{r}$ for the HV calculation based on the worst value of the respective objective that is typically seen on a Pareto front for that task.

We evaluate the influence of priors that are defined by a domain expert and represent a typically chosen trade-off between task performance and safety or impact. Furthermore, we evaluate the impact of misleading priors, which are purposely designed to not adequately solve the task. In practice, the operator prior puts high density on regions of the search space that are believed to yield Pareto-optimal policies, whereas the misleading prior puts very high density on an outer edge of the search space - a choice a reasonable operator would likely not make.

\subsection{Learning in Simulation}
As suggested in~\cite{mayr21iros} and~\cite{mayr22iros}, we learn the tasks in simulation based on a digital twin of the experimental setup.
To have a performance reference, we compute two performance baselines for each of the three tasks where we learn the skill parameters through (1) random search 
and (2) BO with no priors. We repeat every experiment configuration 20 times to account for noise.

\subsubsection{Peg Insertion Task}
The goal of the peg insertion task is to insert a peg into a hole with a \SI{1.5}{\milli\meter} larger radius. The setup imitates a piston insertion into the engine shown in Fig.~\ref{fig:exp-setup}. The configuration of such an insertion does not allow to tilt the piston. Therefore, the insertion strategy is to hold the object upright and to perform an Archimedes spiral as a search motion\insertref{compliance-based robot peg-in-hole assembly strategy without force feedback}. The realization of a spiral is defined by the path velocity of the reference point, the pitch (i.e. the distance between lines in the spiral) and the maximal radius. Furthermore, the insertion skill sets a downward force that is applied by the arm while searching.

As in~\cite{mayr22iros}, this task has two objectives: 1) the performance of the insertion which is assessed with the distance of the peg to the hole as well as a success reward if the peg is inserted by more than \SI{0.01}{\meter} and 2) an integral over the commanded force while searching.

In order to learn a robust solution, each candidate parameter set is evaluated 7 times: (1) each execution randomly selects one out of the five start positions for the robot arm, (2) in simulation the hole is translated horizontally by a Gaussian offset with a standard deviation of \SI{7}{\milli\meter}. 
See~\cite{mayr22iros} for more details about the experiment.

\begin{figure*}[tpb]
	\centering
	{
		\setlength{\fboxrule}{0pt}
		\framebox{\parbox{0.98\textwidth}{
		\centering

		\includegraphics[width=\textwidth]{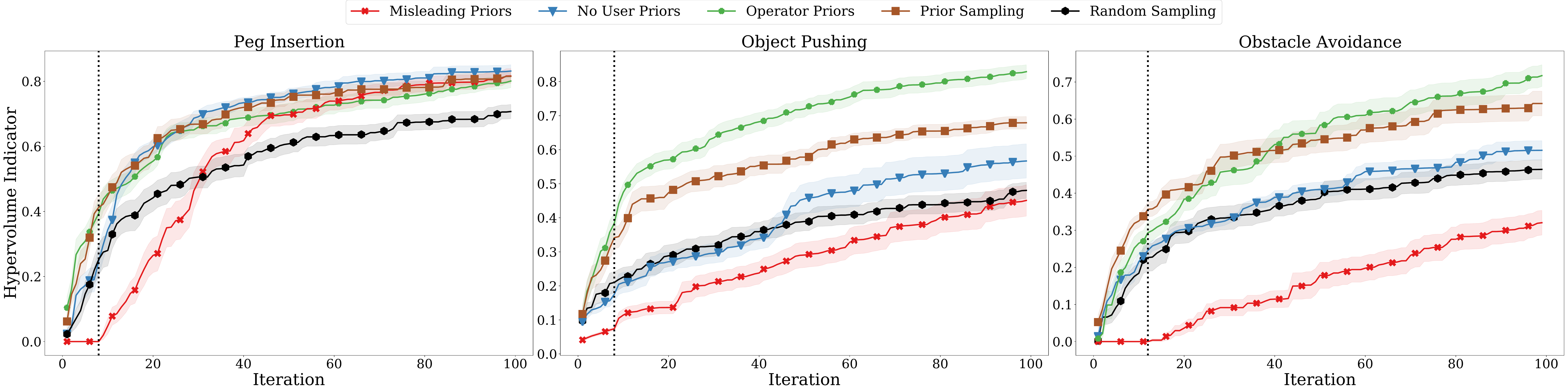}
	
	\caption{The learning progress of the peg insertion, the object pushing and the obstacle avoidance tasks in simulation. The dashed line denotes the end of the DoE phase and the shaded regions are the standard error of the mean. BO with operator priors improves substantially on both BO without priors and prior sampling for two tasks. For the less difficult peg insertion tasks, all approaches but random sampling achieve comparable performance.}
	\label{fig:exp-sim}
	}}}
\end{figure*}

\subsubsection{Object Pushing Task}
This task requires to push an object with an uneven weight distribution from a start location (shown in Fig.~\ref{fig:exp-setup}) to the corner between the block and the metal fixture. The pushing is done with a square peg that is \SI{0.07}{\meter} wide. The parametric push skill first moves the end effector to a location above the object, before it lowers it and performs a Cartesian linear motion towards the goal. The start and goal location of the push movement can be altered in both horizontal directions. This allows for learning of a push motion that starts from a location at the side of the object, and which implicitly takes the center of mass into consideration.
Every parameter set is evaluated 7 times where (1) the  start position is randomly selected from a set of 4 initial positions and (2) the position of the object and the target are slightly perturbed by Gaussian offsets to avoid overfitting.

The performance of the task is assessed by the distance of the end effector to the target pose and by the position and orientation error of the object with respect to the target pose. The other objective assesses the total amount of force that the robot arm applies on the environment by integrating over the error between the actual pose of the end effector and the reference pose. See~\cite{mayr22iros} for a more detailed description.
% &

%
\subsubsection{Obstacle Avoidance Task}
The goal of the task is to find a policy that uses parametric movement skills to avoid a static obstacle in the workspace. The structure of the skill is pre-defined and set up so that an obstacle can be passed from above.  As shown in Fig.~\ref{fig:learnable}, the end effector starts at the point \(s\) and moves towards the goal
\((g_1)\) until it is above the parametric threshold \(p_1\) in \(z\)-direction. When above \(p_1\), the reference point will move towards the goal \((g_2)\) until the threshold \(p_2\) in \(y\)-direction is reached. Then, the motion towards the point \(g_3\) is started. The learnable parameters include the thresholds \(p_1\) and \(p_2\) as well as the \(y\) and \(z\) coordinates of the parametric goal points \((g_1)\) and \((g_2)\). See~\cite{mayr21iros} for additional details.

This task uses a positive reward that evaluates the distance between the end effector and the goal position. Furthermore, there is a fixed reward when reaching the goal. The safety objective evaluates the distance between the end effector and the object and the table.

\subsubsection{Results}

The experimental results are summarized in Fig.~\ref{fig:exp-sim}. In the peg insertion task BO without priors (blue) performed equally well than with the operator priors (green). This task also allowed for a quick recovery from misleading priors (red), indicating that it is easier to learn than the other tasks.
In the other two tasks operator priors (green) greatly improved the learning speed and learning results. Operator priors vastly outperformed the baselines, as it yielded an increase in final HV of about 40\% over BO with no priors. At the same time, less than 40\% of the iterations were needed to achieve the final performance of the baselines. The tasks also showed that the usage of deliberately misleading priors can generally hamper the learning performance.
To provide an additional indication of the performance of the operator priors for a specific task, we performed random sampling in the space that is defined by these priors (brown). While priors sampling performs equally well in the peg insertion task it shows significantly worse performance in the other two tasks.

\subsection{Learning with the Real system}
While learning in simulation has several advantages, it also has limitations: Especially contact-rich tasks require an accurate model of the robot system and the workstation to allow the learning policies to transfer to the real system. This can be difficult to achieve and to maintain. Therefore, we learn the peg insertion task and the obstacle avoidance task directly on the real system.

We learn using the same operator priors as in simulation. As a baseline we use BO without priors. In addition, we also use the Pareto-optimal points that were obtained by learning in simulation as a multimodal prior when learning on the real system as an application of transfer learning. This can be particularly interesting, because it does not necessarily require operators to be able to specify priors. Moreover, the approach allows further refinement on the real system in case the task was not accurately modeled in simulation.

For each of the tasks, we do four repetitions of each configuration to reduce measurement noise. When using the Pareto-optimal points from simulation for a learning process on the real system, we use the points of a single run in simulation that applied BO without priors. This means that the operator never needed to explicitly state any priors.

\subsubsection{Peg Insertion Task}
When learning this task on the real system, we utilize the same five start positions as above when learning in simulation. Every parameter set is evaluated three times, randomly selecting one of the start positions for each run.
\begin{figure}[tpb]
	{
		\setlength{\fboxrule}{0pt}
		\framebox{\parbox{3in}{
		\includegraphics[width=0.85\columnwidth]{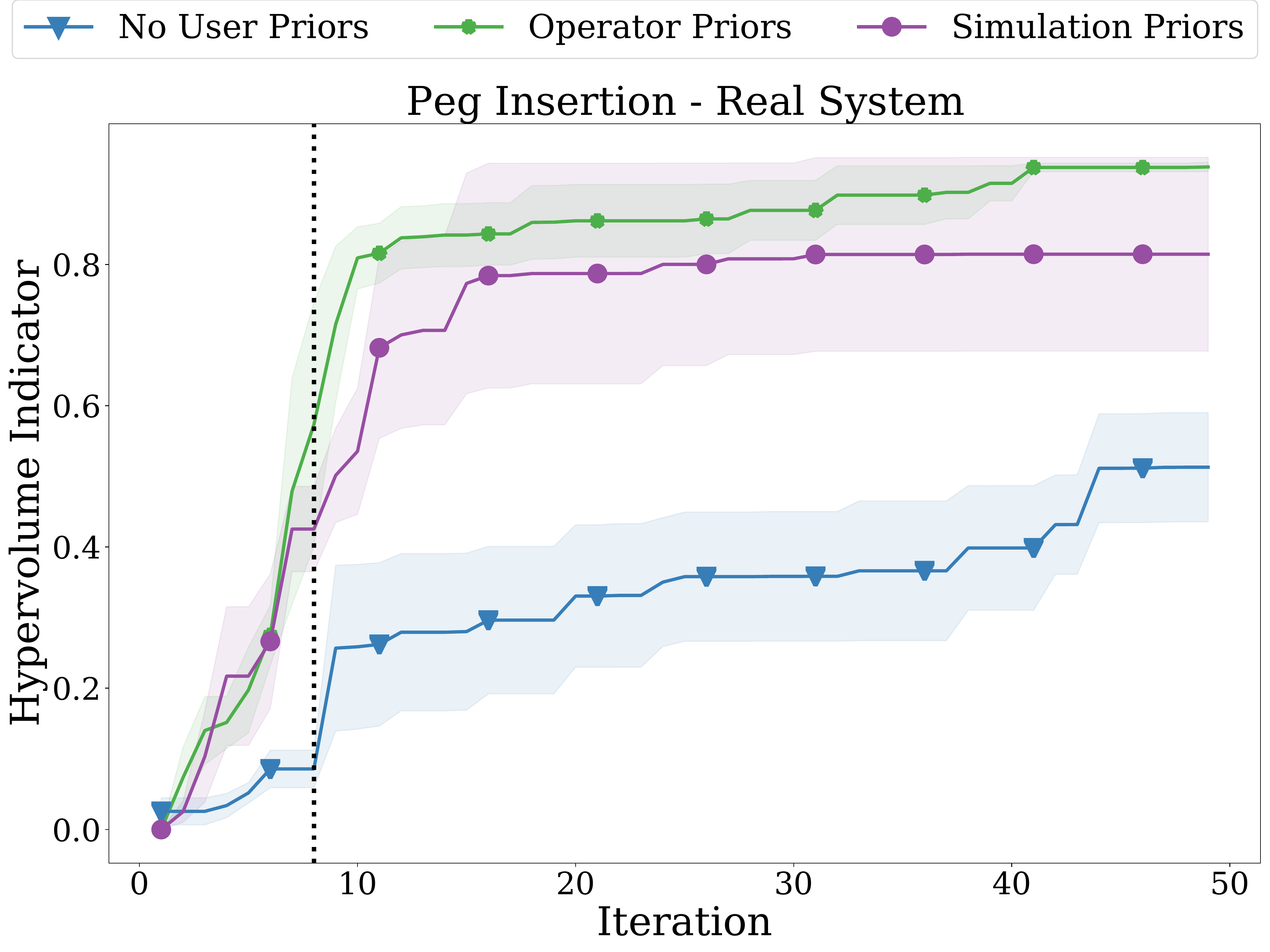}}}
	}
	\caption{The learning progress of the peg insertion task on the real robot system. The dashed line denotes the end of the DoE phase and the shaded regions are the standard error of the mean. Both operator priors and simulation priors yield substantial performance gains over BO without priors.}
	\label{fig:exp-peg-real}
\end{figure}
\subsubsection{Obstacle Avoidance Task}
Since learning this task on the real robot system can result in collisions with the object, the engine in Fig.~\ref{fig:exp-setup} is replaced by an object that avoids damages to the robot. Furthermore, since successful policies do not interact with the environment in this task, every parameter set is evaluated only once.

\begin{figure}[tpb]
	{
		\setlength{\fboxrule}{0pt}
		\framebox{\parbox{3in}{
		\includegraphics[width=0.85\columnwidth]{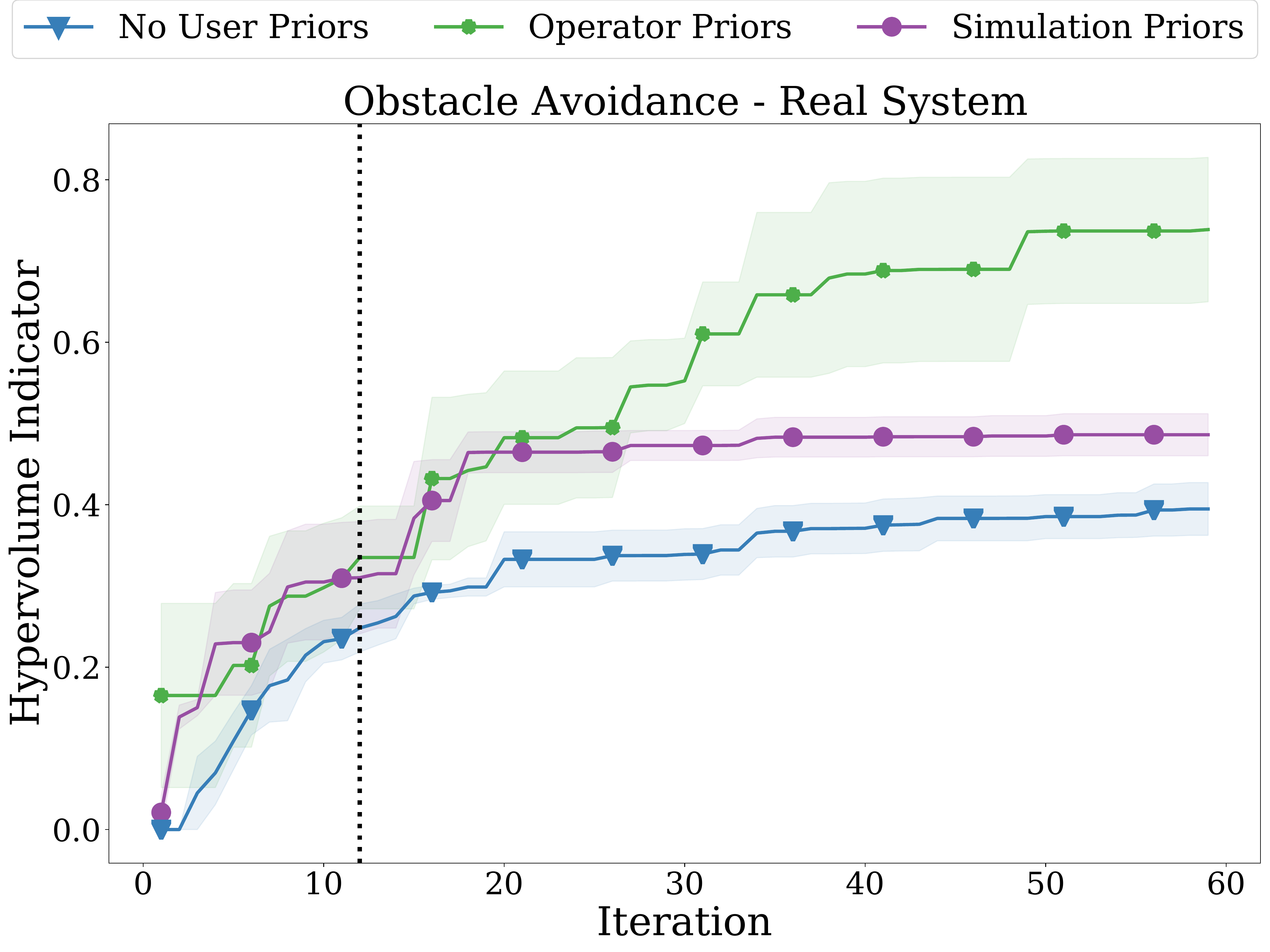}}}
	}
	\caption{The learning progress of the obstacle avoidance task on the real robot system. The dashed line denotes the end of the DoE phase and the shaded regions are the standard error of the mean. Both operator priors and simulation priors yield substantial performance gains over BO without priors.}
	\label{fig:exp-obstacle-real}
\end{figure}

\subsubsection{Results}
The experimental results are summarized in Fig. \ref{fig:exp-peg-real} and \ref{fig:exp-obstacle-real}. They demonstrate again that well-placed operator priors could accelerate the learning and yielded to better learning results. In both tasks the priors derived from simulation results perform equally well initially, but can eventually be outperformed by the operator priors. However, since the simulation priors do not need to be defined explicitly, they are particularly interesting for new tasks or less experienced operators. In both tasks it takes less than 30\% of the iterations to achieve the same performance as BO without priors.
When learning the peg insertion task, the average force applied by the robot was \SI{32}{\percent} lower when using priors from simulation and \SI{47}{\percent} lower with operator priors compared to a policy search without priors.
In the obstacle avoidance task, the average amount of required interferences by robot operators due to forceful collisions with the object was much lower when learning with operator (\num{10.75(443)}) and simulation priors (\num{4.5(8)}) than with BO without priors (\num{24.25(443)}). This indicates that even the safety of a policy search can be increased by incorporating well-chosen priors.

\section{CONCLUSIONS}
We evaluated the influence of prior beliefs about the location of good candidate solutions when learning several industrial robot tasks. Since the parameters of interpretable robot skills often have a concrete meaning, they offer a natural opportunity for robot operators to incorporate their knowledge and experiences into the learning process. We have shown that expert operator priors can substantially speed up the search and yield higher performing policies, and seldom harm the performance. We have also demonstrated how using results from learning in simulation as priors can automate the prior design, and how this choice accelerates the learning of the same task on the real robot system. Lastly, we have highlighted the risk and potential performance loss associated with specifying a drastically incorrect prior. 

We believe that the usage of priors for robot tasks learning in a combination with skills and the knowledge integration is a promising direction to achieve intelligent  robot systems that can quickly learn to adapt. Moreover, the usage of priors can ease the adaption of RL in industrial robot tasks by providing operators an intuitive tool to guide a learning process.
We are planning to look more into the transfer of knowledge between different tasks and robot configurations. Furthermore, multi-fidelity learning could combine a small amount of executions on the real system with learning in simulation to allow for a quicker and safer adjustment to new tasks.
%\addtolength{\textheight}{-12cm}   % This command serves to balance the column lengths
                                  % on the last page of the document manually. It shortens
                                  % the textheight of the last page by a suitable amount.
                                  % This command does not take effect until the next page
                                  % so it should come on the page before the last. Make
                                  % sure that you do not shorten the textheight too much.

%%%%%%%%%%%%%%%%%%%%%%%%%%%%%%%%%%%%%%%%%%%%%%%%%%%%%%%%%%%%%%%%%%%%%%%%%%%%%%%%

%%%%%%%%%%%%%%%%%%%%%%%%%%%%%%%%%%%%%%%%%%%%%%%%%%%%%%%%%%%%%%%%%%%%%%%%%%%%%%%%

%%%%%%%%%%%%%%%%%%%%%%%%%%%%%%%%%%%%%%%%%%%%%%%%%%%%%%%%%%%%%%%%%%%%%%%%%%%%%%%%
\section*{Appendix}
The implementation as well as additional information are available at: \small{\url{https://github.com/matthias-mayr/SkiREIL}}

\section*{Acknowledgement}
This work was partially supported by the Wallenberg AI, Autonomous Systems and Software Program (WASP) funded by Knut and Alice Wallenberg Foundation.
This research was also supported in part by affiliate members and other supporters of the Stanford DAWN project — Ant Financial, Facebook, Google, InfoSys, Teradata, NEC, and VMware.

“© 2022 IEEE.  Personal use of this material is permitted.  Permission from IEEE must be obtained for all other uses, in any current or future media, including reprinting/republishing this material for advertising or promotional purposes, creating new collective works, for resale or redistribution to servers or lists, or reuse of any copyrighted component of this work in other works.”

\addtolength{\textheight}{-7cm}
%%%%%%%%%%%%%%%%%%%%%%%%%%%%%%%%%%%%%%%%%%%%%%%%%%%%%%%%%%%%%%%%%%%%%%%%%%%%%%%%
\bibliography{bib/2022-CASE}
\bibliographystyle{bib/IEEEtran}

\end{document}